\documentclass{article}

    \PassOptionsToPackage{numbers, compress}{natbib}
    \usepackage[preprint]{neurips_2023}

\usepackage[utf8]{inputenc} % allow utf-8 input
\usepackage[T1]{fontenc}    % use 8-bit T1 fonts
\usepackage{hyperref}       % hyperlinks
\usepackage{url}            % simple URL typesetting
\usepackage{booktabs}       % professional-quality tables
\usepackage{amsfonts}       % blackboard math symbols
\usepackage{nicefrac}       % compact symbols for 1/2, etc.
\usepackage{microtype}      % microtypography
\usepackage{xcolor}         % colors

\usepackage{algorithm}
\usepackage{algorithmic}
\usepackage{enumitem}
\usepackage{graphicx}
\usepackage{amssymb}
\usepackage{mathrsfs}
\usepackage{amsmath}
\usepackage{mathtools}
\usepackage{subfigure}
\usepackage{graphicx}
\usepackage{wrapfig}
\usepackage{multirow}
\usepackage{natbib}

\newtheorem{definition}{Definition}

\newcommand\algo{\emph{MBeDED }}

\title{Marginal Benefit and Diversity Induced Unsupervised Environment Design}

\author{%
  Dexun Li\thanks{This is the initial version of our submission} \\
  Singapore Management University\\
  \texttt{dexunli.2019@phdcs.smu.edu.sg} \\
  \And
  Wenjun Li \\
  Singapore Management University \\
  \texttt{wjli.2020@phdcs.smu.edu.sg} \\
  \AND
  Pradeep Varakantham \\
  Singapore Management University \\
  \texttt{pradeepv@smu.edu.sg} \\
}

\begin{document}

\maketitle

\begin{abstract}
Recent work on designing an appropriate distribution of environments has shown promise for training effective generally capable agents. Its success is partly because of a form of adaptive curriculum learning that generates environment instances (or levels) at the frontier of the agent's capabilities. However, such an environment design framework often struggles to find effective levels in challenging design spaces and requires costly interactions with the environment. In this paper, we aim to introduce diversity in the Unsupervised Environment Design (UED) framework. 
Specifically, we propose a task-agnostic method to identify observed/hidden states that are representative of a given level. The outcome of this method is then utilized to characterize the diversity between two levels, which as we show can be crucial to effective performance. In addition, to improve sampling efficiency, we incorporate the self-play technique that allows the environment generator to automatically generate environments that are of great benefit to the training agent. Quantitatively, our approach, \emph{Div}ersity-induced Environment Design via \emph{S}elf-\emph{P}lay (\algo), shows compelling performance over existing methods.

\end{abstract}

\section{Introduction}

The advances in Reinforcement Learning (RL) have recorded great success in a variety of applications, such as game playing~\cite{mnih2015human,silver2016mastering} and robot control~\cite{levine2016end,akkaya2019solving}. However, using RL to train agents with general capabilities remains a major challenge. This is because RL utilizes stochastic exploration, which requires collecting millions of samples, and interacting with the environment is very expensive and time-consuming. Therefore, it is essential to improve sample efficiency and build effective exploration.

Training an agent by specifying a distribution of tasks or environments is one promising approach to improve the overall capabilities of RL agents~\cite{dennis2020emergent,jiang2021prioritized,parker2022evolving,li2023effective, tio2023transferable}. This type of approach automatically generates challenging environments, giving rise to a curriculum that can constantly promote the frontier of the agent's capabilities. For example, we can control a set of parameters that may correspond to the position of an obstacle in a maze environment or the coefficient of friction in a robot simulator. Then each set of parameters will produce an instance of the environment, which we call a level~\footnote{We use level and environment interchangeably in the paper}. By adapting the training distribution over the parameters of environments, adaptive curricula have been shown to produce more robust agents in fewer training steps~\cite{portelas2020teacher,jiang2021prioritized}.

Typically, Unsupervised Environment Design (UED,~\cite{dennis2020emergent}) introduces a self-supervised RL paradigm where they formalize the problem of generating a distribution over environments by considering an environment policy (teacher) to curate such an environmental distribution to best support the continued training of a particular agent policy (student). Here, the teacher is co-evolved with the student and is trained to produce difficult but solvable environments that maximize the defined object of regret. \citet{dennis2020emergent} showed that this would lead to a form of adaptive curriculum learning and that the student's policy is the minimum regret strategy when the multi-agent system reaches Nash equilibrium. However, training such a teacher-student framework remains a challenge because the teacher's policy update relies on the regret value, which is approximated by the difference between the expected payoffs of the protagonist and the antagonist, resulting in costly interactions between both student agents and the environment. In contrast, another line of work, Prioritized Levell Replay (PLR,~\cite{jiang2021prioritized}), embodies an alternative form of dynamic curriculum learning that does not assume control of level generation; but instead, PLR will selectively revisit/replay existing randomly generated levels with probability proportional to their corresponding regret values. This type of approach would suffer from the costly randomly generated level, it would not be able to exploit any previously discovered level structures, and its performance is highly influenced by the magnitude of possible environment parameters. 

Notably, exploration is a key problem in reinforcement learning, since agents can only learn from data they acquire in the environment. In this paper, we attempt to build an effective exploration and regret-based curriculum in a sample efficient manner. We harness the advantages from both PAIRED and PLR and introduce a novel diversity-driven approach, named \emph{Div}ersity-induced Environment Design via \emph{S}elf-\emph{P}lay (\algo). \algo maintains a set of diverse and useful environments that allow data to be collected in an informative manner. Specifically,  to encourage diversity, we propose a task-agnostic observed/hidden states representative method to address the problems of measuring the diversity score between two environments. More importantly, \algo is able to automatically generate environments at the frontier of the student agent's capabilities while avoiding the costly interaction with the environment by incorporating the technique of Self-Play. By combining diversity and learning potential, our algorithm enables us to validate the effectiveness of the current training level, achieving a dynamic balance between diversity and marginal benefit, effectively catalyzing the general ability of the agent.

% Notably, exploration is a key problem in reinforcement learning, since agents can only learn from data they acquire in the environment. With that in mind, maintaining a set of diverse environments is an attractive method, as it allows data be collected with an informative manner. In this paper, we present a novel diversity-driven exploration strategy which is based on the framework of PAIRED, in order to encourage diversity, we propose a sampling-based state representation method to tackle the problems of measuring the diversity objective between two environments, The distance measure evaluates diversity score in the replay buffer, and we keep the environment replay buffer to maximize the diversity information between agent's trajectories. This objective enables agent to distinguish two different environments that have value of training. Structurally, we further propose a complementary approach that relies on the delayed self-policy. In particular, instead of liking PAIRED that we need to train a protagonist and antagonist simultaneously, if we keep track of the agent policy, we can compute the marginal improvement of the benefit through training on a level. This enables us to validate the effectiveness of the current training level. Combining these novelties achieves a dynamic balance between the diversity and marginal benefit, efficiently catalyzing the general capability of the agent.

Overall, our contributions are three-fold as follows:
\begin{itemize}[leftmargin=*]
    \item We propose a diversity-driven approach that maintains a diverse level buffer with informative and useful environments to enhance effective exploration.
    \item Taking advantage of the self-play technique, we avoid costly training of both protagonist and antagonist, and our algorithm learns to automatically generate environments in a sample efficient way.
    \item We empirically demonstrate that \algo outperforms existing methods while achieving a speedup compared to other methods for automatic environment generation.
\end{itemize}

\section{Preliminaries}
\subsection{Underspecified Partially Observable Markov Decision Process}
A Markov Decision Process (MDP) is defined as a tuple $<\mathcal{S}, \mathcal{A}, P, R, \gamma, T>$. Here $\mathcal{S}$ and $\mathcal{A}$ stand for the set of states and actions respectively, $P(s_{t+1}|s_t)$ is the probability that the state transitions from $s_t\in \mathcal{S}$ to $s_{t+1}\in \mathcal{S}$ given action $a_t$, and $r_t = R(s_t, a_t, s_{t+1})$ is a reward obtained by the agent transitioning from $s_t$ to $s_{t+1}$ when taking the action $a_t$. $\gamma$ is the discount factor, and $T$ is the horizon. Given an MDP, the goal of an RL agent is to learn the policy $\pi$ such that the cumulative discounted reward, i.e., $\mathbb{E}[\sum_t=0^T\gamma^T r_t]$, is maximized.

However, the MDP framework is not feasible to describe most of the real-world applications as the agent is unable to directly observe the underlying whole state. Furthermore, the UED problem requires using an underspecified environment to produce a distribution over fully specified environments, resulting in the state space and transition probability change according to the specified environment. To model the fully specified environments with Partially Observable Markov Decision Process (POMDP) in the UED framework, \citet{dennis2020emergent} first introduce the Underspecified POMDP (UPOMDP) as a tuple $\mathcal{M} = < \mathcal{A}, \mathcal{O}, \Theta, \mathcal{S}^{\mathcal{M}}, P^{\mathcal{M}}, \mathcal{I}^{\mathcal{M}}, R^{\mathcal{M}}, \gamma, T>$. Here $\mathcal{O}$ is a set of observations, and $\mathcal{I}^{\mathcal{M}}: \mathcal{S}\rightarrow\mathcal{O}$ is the observation function. Compared to POMDP, there is an addition of $\Theta$ in UPOMDP, which represents the free parameters of the environment. Those parameters are incorporated into the transition function and reward function. Following~\cite{jiang2021replay}, we define a level $\mathcal{M}_{\theta}$ as an environment resulting from a fixed $\theta \in \Theta$. The objective of the RL agent policy $\pi$ is to maximize the value $V^\theta(\pi)$ in $\mathcal{M}_{\theta}$, which is $V^\theta(\pi)=\mathbb{E}[\sum_{t=0}^T \gamma^t r_t]$, and $r_t$ are the rewards collected by $\pi$ in $\mathcal{M}_{\theta}$.

\subsection{Existing Methods}
% \citet{tobin2017domain} first bridge the "reality gap" between simulated robotics and real robots. They randomize the simulator so that the models are exposed to a wide variety of environments during training. They argue that if the variability of the simulation is large enough, models trained in the simulation will generalize to the real world with no additional training. In contrast, Unsupervised Environment Design (UED) is an alternative paradigm. 
\citet{dennis2020emergent} first formalize the UED and introduce the Protagonist Antagonist Induced Regret Environment Design (PAIRED) algorithm, which is a three-agent game: the protagonist $\pi^P$, the antagonist $\pi^A$ and the environment generator $\mathcal{G}$. 
The environment generator $\mathcal{G}$ learns to control the distribution of environmental parameters by maximizing regret, which is approximated by the difference between the cumulative reward obtained by the protagonist and the adversary agent under a fixed environment $theta$, respectively:
\begin{equation}
     \text{REGRET}^{\theta}(\pi^P,\pi^A) = V^{\theta}(\pi^A)-V^{\theta}(\pi^P)
\end{equation}
The protagonist and antagonist are both trained to maximize their own cumulative reward in current environments. Note that the environment generator (teacher) is discouraged from generating levels that can not be solved because they will have a maximum regret of 0. This teacher-student framework co-evolves teacher and student policies, creating a form of adaptive curriculum learning in which the teacher constantly creates an emergent class of levels that get progressively more difficult along the borderline of the protagonist's ability, allowing agents to learn a good policy that enables zero-shot transfer. More specifically, if $\Pi$ is the strategy set of the protagonist and antagonist, and $\Theta$ is the strategy set of the teacher, then if the learning process reaches a Nash equilibrium, the resulting student policy $\pi$ provably converges to a minimax regret policy, defined as 
\begin{equation}
    \pi = \underset{\pi^P\in \Pi}{\arg \min} \lbrace \underset{\theta, \pi^A\in \Theta, \Pi}{\max} \{
 \text{REGRET}^{{\theta}}(\pi^P,\pi^A) \}  \rbrace
\end{equation}
However, this teacher-student framework struggles with the building of an efficient generator (teacher), as it requires expensive interactions with the environment to collect millions of samples to train Protagonist and Antagonist agents separately.
As an alternative regret-based UED approach, \citet{jiang2021prioritized} propose Prioritized Level Replay (PLR), where a student policy is challenged by two co-evolving teachers, Generator and Curator. In PLR, Generator randomly creates new environments and Curator prioritizes the replay probability for each environment based on the estimated learning potential in each environment. By adapting the sampling of the previously encountered levels to train, PLR is an active learning strategy that improves sample efficiency and generalization. In addition, PLR uses the \emph{positive value loss} to approximate the regret compared to the generative but slow adaptive PAIRED. Specifically, regret is approximated by:
\begin{equation}\label{eq:gae}
    F_{gae}(\theta) = \frac{1}{T}\sum_{t=0}^T \max\{  \sum_{k=t}^T(\gamma \lambda)^{k-t} \delta_k ,0 \}
\end{equation}
where $\lambda$ and $\gamma$ are the Generalized Advantage Estimation (GAE,~\cite{schulman2015high}) and MDP discount factor respectively. $\delta$ is the TD-error at timestep $t$. The agent trained by PLR shows good generalization ability  in terms of empirical results. However, PLR is still limited as it is unable to exploit any previously discovered level structure and can only curate randomly sampled levels; moreover, the random search will be heavily affected by the high-dimensional design space, making it highly unlikely to sample levels at the frontier of the agent's current capabilities.

\section{Algorithm}
In this section, we propose a novel generic diversity-driven UED appraoch that can improve the sampling efficiency and generality of RL agents. The overall algorithm is summarized in Algorithm~\ref{alg:alg}.

\begin{algorithm}[t]
   \caption{\algo}
   \label{alg:alg}
\begin{algorithmic}[1]
   \STATE {\bfseries Input:} Level buffer $\Lambda$ of size $K$, replay probability $p$. 
   Initialize policy Alice $\pi^A$, Bob $\pi^B$, level generator $\mathcal{G}$;
   % \STATE \# Sample $K\cdot \rho$ initial levels/
   % \STATE \# Main training Loop
   \WHILE{Not converged}
   \STATE Sample a replay decision, $\epsilon\sim U[0,1]$
   \IF{$\epsilon  \leq p$}
    % \STATE \# Evaluate DR levels and also update
   % \STATE Sample level $\theta_{ep}$ from level generator
   \STATE Use $\mathcal{G}$ to generate new environment parameters $\theta$, and create POMDP $\mathcal{M}_{\theta}$
   \STATE Collect Alice's and Bob's trajectories $\tau^A$ and $\tau^B$ in $\mathcal{M}_{\theta}$, and compute $V^{\theta}(\pi^A) = \sum_{t=0}^{T}\gamma^t r_t$ and $V^{\theta}(\pi^B) = \sum_{t=0}^{T}\gamma^t r_t$, respectively
   \STATE Compute the REGRET using Equation~\ref{eq:regret}
   \STATE Update Bob's policy $\pi^B$ by letting $\pi^B=\pi^A$
   \STATE Train Alice's policy $\pi^A$ to maximize $V^{\theta}(\pi^A)$
   \STATE Train $\mathcal{G}$ with RL update and reward $R=\text{REGRET}$
   \STATE Get observed state representative from trajectory $\tau^A$ according to Equation~\ref{eq:rep_step}
   \STATE If its state-aware diversity score is lower than the highest one in the buffer, replace that one by adding the new level $\theta$ to $\Lambda$ and update the replay probability $P_{replay}$ according to Equation~\ref{eq:replay_prob}
   \ELSE
   \STATE Sample level $\theta$ from $\Lambda$ according to $P_{replay}$, and create POMDP $\mathcal{M}_{\theta}$
   \STATE Collect Alice's and Bob's trajectories $\tau^A$ and $\tau^B$ in $\mathcal{M}_{\theta}$, and compute $V^{\theta}(\pi^A) = \sum_{t=0}^{T}\gamma^t r_t$ and $V^{\theta}(\pi^B) = \sum_{t=0}^{T}\gamma^t r_t$, respectively
    \STATE Update Bob's policy $\pi^B$ by letting $\pi^B=\pi^A$
   \STATE Train Alice's policy $\pi^A$ to maximize $V^{\theta}(\pi^A)$
   \STATE Update the observed state representative for level $\theta$ according to Equation~\ref{eq:rep_step}
   \STATE Update the replay probability $P_{replay}$ according to Equation~\ref{eq:replay_prob}
   \ENDIF 
   % \STATE Update policy $\pi_2$ by letting $\pi_2=\pi_1$
   % \STATE Update policy $\pi_1$ with rewards $R(\tau_1)$
   %  \STATE Update environment generator using advantage score
   % \STATE Let $\kappa = \gamma*\kappa$
   % \STATE \# Using domain randomization to produce \textcolor{blue}{$\theta^{\prime}_{ep-1}$} based on \textcolor{red}{$\theta_{ep-1}$}
   % \STATE \# Collect $\pi_1$ and $\pi_2$'s trajectories $\tau_1$ and $\tau_2$ on \textcolor{blue}{$\theta^{\prime}_{ep-1}$}, with a stop-gradient
   % \STATE \# Compute the score \textcolor{blue}{$S^\prime$}, using equation~\ref{eq:advantage}
   % \STATE \# Add $\theta^\prime$ to $\Lambda$ if score \textcolor{blue}{$S^\prime$} is bigger than \textcolor{red}{$S$}
   \ENDWHILE
\end{algorithmic}
\end{algorithm}

\begin{figure*}[t]
    \centering
    \includegraphics[width=0.8\linewidth]{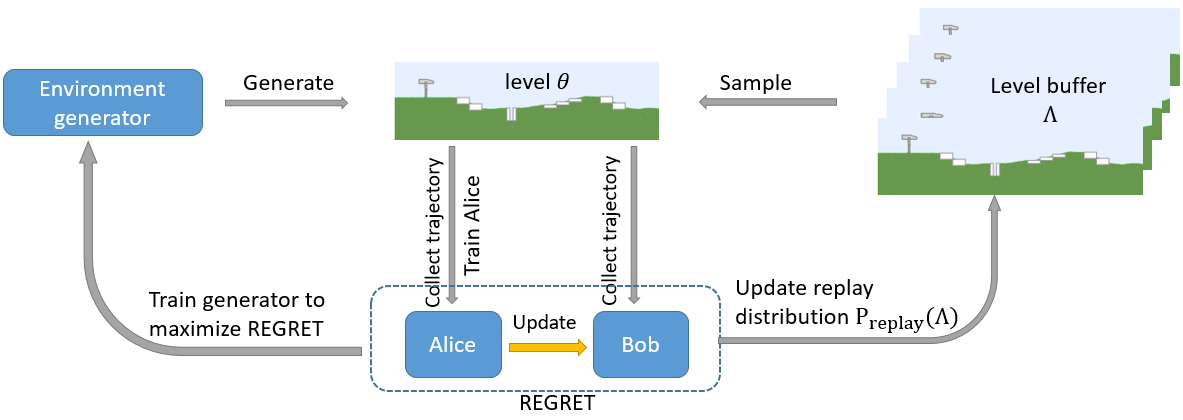}
    \caption{The overall framework of \algo for Unsupervised Environment Design }
    \label{fig:framework}
\end{figure*}

\subsection{Regret-based Generator with Self-Play}
We first introduce how to encourage the environment generator (teacher) to automatically generate levels that are at the frontier of an agent's capabilities by designing a marginal improvement-based regret through Self-Play. We consider the teacher-student framework with a single student agent, but we allow it to have two separate "minds": Alice ($\pi^A$) and Bob ($\pi^B$). They have the same objective and Alice shares the old policy with Bob. More specifically, during the self-play phase, both Alice and Bob will collect several trajectories within the current level $\theta$ (Line 6 and 15 in Algorithm~\ref{alg:alg}), and then we compute the approximate regret value by the difference between the rewards they received (Line 7 in Algorithm~\ref{alg:alg}). Formally, regret is approximated as:
\begin{equation}\label{eq:regret}
\begin{aligned}
        \text{REGRET}^{\theta}(\pi^A,\pi^B) &\approx V^{\theta}(\pi^A)-V^{\theta}(\pi^B)\\
        &= \mathbb{E}_{\tau^A}[V^{\theta}(\pi^A)] - \mathbb{E}_{\tau^B}[V^{\theta}(\pi^B)]
\end{aligned}
\end{equation}
Note that there are alternative methods to compute regret, i.e., using the difference of the scoring function based on the average magnitude of the Generalized Advantage Estimate (GAE;~\cite{schulman2015high}) over each of $T$ time steps. In this work, we used the difference between Bob and Alice's expected cumulative returns to approximate regret because it is straightforward and shows encouraging results for building difficult but solvable environments.
% Note that there are alternative methods for computing regret, i.e., using the difference of the scoring function based on the average magnitude of the Generalized Advantage Estimate (GAE;~\cite{schulman2015high}) over each of $T$ time steps in one trajectory from current level:
% \begin{equation}
% \begin{aligned}
%     \text{REGRET}&^{\theta}(\pi^A,\pi^B) \approx \textbf{score}(\tau^A,\pi^A) - \textbf{score}(\tau^B,\pi^B) \\
%     & = \frac{1}{T}\sum_{t=0}^T|\sum_{k=t}^T(\gamma \lambda)^{k-t}\delta_k^B|-\frac{1}{T}\sum_{t=0}^T|\sum_{k=t}^T(\gamma \lambda)^{k-t}\delta_k^B|.
% \end{aligned}
% \end{equation}

% Here $\delta^A = r_t+\gamma V^(s_{t+1}) - V(s_t)$ is the TD-error at time step $t$ (We compare the experiment result of the choice of the regret calculation in Appendix). 

When generating a new environment, the environment generator (teacher) is trained to maximize the approximated regret (Line 10). The key idea is that Bob's play with Alice should help the environment generator understand how the previously discovered structure works and enable it to automatically generate a more challenging level to encourage further robustness and generalization.
Furthermore, Bob will directly copy current Alice's policy (Line 8 and line 16), avoiding the drawback of PAIRED, which requires expensive interactions with the environment to train an optimal antagonist's policy at the current level. Instead, Alice is trained to maximize the corresponding cumulative reward (line 9 and line 17). 
Overall, this self-regulating feedback between Alice and Bob allows the environment generator to establish an adaptive curriculum: new levels that could lead to a dramatic difference in Alice and Bob's behavior are constantly produced, leading to a more robust agent policy.

\begin{figure*}[t]
    \centering
    \includegraphics[width=1.0\linewidth]{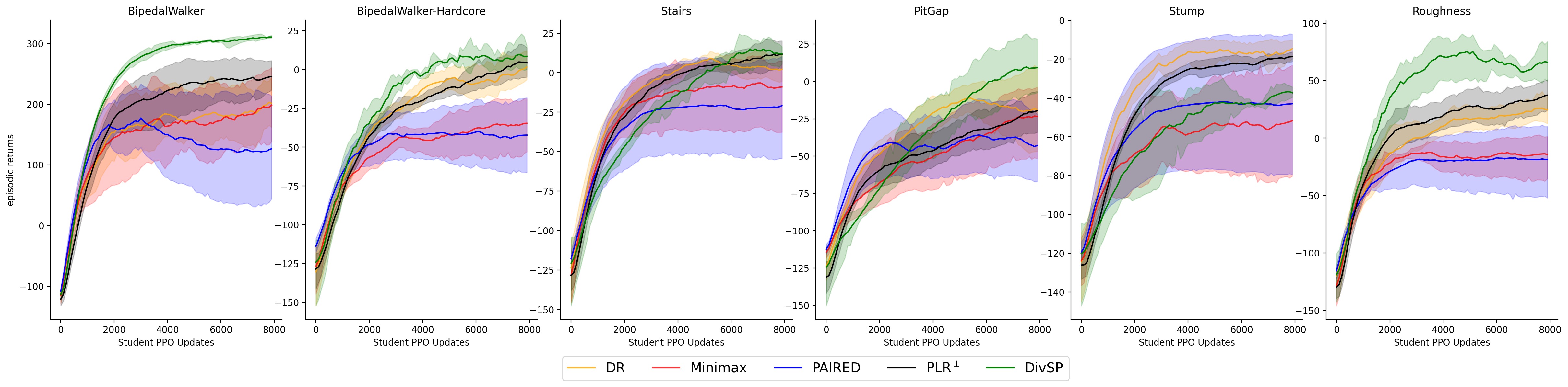}
    \caption{Performance on test environments during training (mean and standard error).}
    \label{fig:walker_result}
\end{figure*}

\subsection{State-Aware Diversity}
The objective of this section is to apply diversity to the UED framework to build effective explorations and improve sample efficiency.

% We now seek to improve the sample efficiency of our approach by introducing diversity to the UED setting and derive a state-aware diversity represent method for training in a meaximally diverse population of environments. sampling the next training level by prioritizing those with higher estimated learning potential when revisited in the future.

Existing algorithms (\cite{jiang2021replay,jiang2021prioritized,parker2022evolving}) consider a setup whereby a level buffer $\Lambda$ is introduced to store the top $K$ visited levels with the highest learning potential, which is estimated by the GAE value of the learning agent over the last episode. However, determining the level buffer by the learning potential alone may result in preserving similar or repeated levels, so it would suffer from low-quality exploration, and the agent will not learn much from training on these repeated collected trajectories. As a result, we present a diversity-driven strategy to maintain a set of diverse environments so that data can be collected in an informative manner. To do so, we first need to specify a \emph{Diversity} measure that can be used to distinguish the diversity between two levels.
We derive a state-aware diversity representative method as follows:

\textbf{From teacher's perspective}: Two levels $\theta_1$ and $\theta_2$ can be said to be different if the distance between the parameters of the environment is large. However, this method is not interpretable because the mapping from the parameters to levels is not linear. For example, in the continuous-control \emph{BipedalWalker} environment, the environment design space is an 8-dimensional indirect encoding representing the intensity of four kinds of obstacles for the student agent. We are unable to capture the stochasticity in the mapping from parameters to the environment. Additionally, the normalization of different parameters is domain-specific, which prevents us from directly using environment parameters to measure the diversity of different levels.

\textbf{From student's perspective}: Intuitively, to encourage the specialty of individual levels in the replay buffer, student agents need to obtain different local observations to highlight levels from others. To achieve this goal, we can sample a set of states that best represent the structure of the environment, and then use these sets of representative states to measure the diversity in the level buffer. 

Measuring diversity among levels requires a function to measure the diversity of the whole population as the volume of the inter-state matrix. Inspired by~\cite{parker2020effective}, we consider a smooth kernel $k$, such that $k(o_1, o_2)\leq 1$ for observed state $o_1, o_2\in\mathcal{O}$. A popular choice of the kernel is cosine similarity, which is defined as follows:
\begin{equation}
    k(o_1, o_2) = \frac{o_1 o_2^\top}{\Vert o_1\Vert \Vert o_2\Vert},
\end{equation}
and we have:
\begin{equation}
    k(o_1, o_2) = 1 \iff o_1 = o_2.
\end{equation}
Note that if we use a Recurrent Neural Network (RNN) to parameterize the student agent, such as the partially observable navigation task domain, we can also replace the observed state with the output of the hidden layer of the RNN agent. With a flexible method of measuring the inter-state diversity at hand, we can give a formal definition of our observed state representative method.

\begin{definition}
    (\textbf{Observed State Representative}) For each environment $\theta$, considering several trajectories induced by the current student policy, we denote $O=\{o_1, \dots, o_m \}$ as the set of observed states of size $m$ collected from the current level, and we can derive a set of observed state $S_{env}$ of size $n$, where $n\ll m$, to best represent the environment $\theta$. Formally, the representative score is defined as:
    \begin{equation}\label{eq:def_rep}
        F_{rep}(S_{env}) = \sum_{o_i \in O} \underset{o_j \in S_{env}}{\max} \{  k(o_i, o_j) \}
    \end{equation}
    and we have $S_{env} = \underset{S_{env} \subset O, |S_{env}|\leq n}{\arg  \max} F_{rep}(S_{env})$
\end{definition}
Intuitively, we use the cosine similarity kernel $k$ to measure how well the selected observed states in the set $S_{env}$ can represent the whole observed states $O$ collected from the current level. A high representative score $F_{rep}(S_{env})$ indicates that each observed state collected from the current level can find a sufficiently similar state in $S_{env}$, such that $S_{env}$ is a good representative of the current level.
However, getting the exact solution of set $S_{env}$ from $O$ is NP-hard, and it is costly since the set $O$ is very large. Motivate by~\citet{li2021claim}, we propose a heuristic way to do so:
\begin{enumerate}
    \item We first randomly sample a set $O^\prime \subset O$ of size $m^\prime$, where $n<m^\prime\ll m$;
    \item Then we use a greedy algorithm to pick the top $n$ observed states from the set $O^\prime$. we start by taking $S_{env}$ as an empty set and at each step, we will add the observed state $o$ that maximizes the marginal gain, where the marginal gain $F_{rep}(o|S_{env})$ is defined as the difference of adding observed state $o$ into $S_{env}$:
    \begin{equation}\label{eq:rep_step}
        F_{rep}(o|S_{env}) = F_{rep}(o \cup S_{env})- F_{rep}(S_{env})
    \end{equation}
\end{enumerate}
Because $F_{rep}(S_{env})$ defined in Equation~\ref{eq:def_rep} is a submodular function, the greedy algorithm can provide a solution with an approximation factor $1-\frac{1}{e}$~\cite{nemhauser1978analysis}. 

We now consider the diversity of two levels. The intuition for doing so is that a diverse level buffer is more informative and thus contributes to effective exploration and sampling efficiency.
The formal definition of measuring the state-aware diversity among levels is as follows.

\begin{definition}
(\textbf{State-Aware Diversity}) Consider the level buffer $\Lambda=\{ \theta_1,\dots,\theta_K \}$ and the newly generated level $\theta_{new}$, each of which has its corresponding observed state representative $S_{env}$. For the newly generated level or any level in the level buffer $\theta_{i} \subset \{ \theta_{new} \} \cup \Lambda  $, we can compute its state-aware diversity score $F_{div}(,)$ among other levels as follows:
\begin{equation}\label{eq:def_div}
    F_{div}(\theta_{i},\{ \theta_{new} \} \cup \Lambda \setminus \{ \theta_i\} ) = \sum_{o_i\in S_{env}}\underset{o_j\in S_{env}^\prime}{\max} \{ k(o_1, o_2) \},
\end{equation}
where $S_{env}$ is the observed state representative for level $\theta_i$, and $S_{env}^\prime$ is the set of observed state representative for levels in $\{ \theta_{new} \} \cup \Lambda \setminus \{ \theta_i\} $. Similarly, the state-aware diversity of level buffer is:
    \begin{equation}
            F_{div}(\Lambda,\Lambda) = \sum_{\theta_i\in \Lambda} F_{div}(\theta_i,\Lambda \setminus \{ \theta_i\})
    \end{equation}
\end{definition}

% When there is a newly generated level, we will measure the diversity score of the new level and the levels in the level buffer according to Equation~\ref{eq:def_div}, and then we can decide whether to add the new levels to the buffer to maximize the diversity score of the level buffer (Line 11 and 12 in Algorithm~\ref{alg:alg}).

Finally, at the beginning of each iteration, \algo either generates new levels (with probability $p$, line 4 and 5) or sample a mini-batch of levels in the level buffer to train the agent (line 13 and 14):

\begin{figure*}[t]
    \centering
    \includegraphics[width=0.83\linewidth]{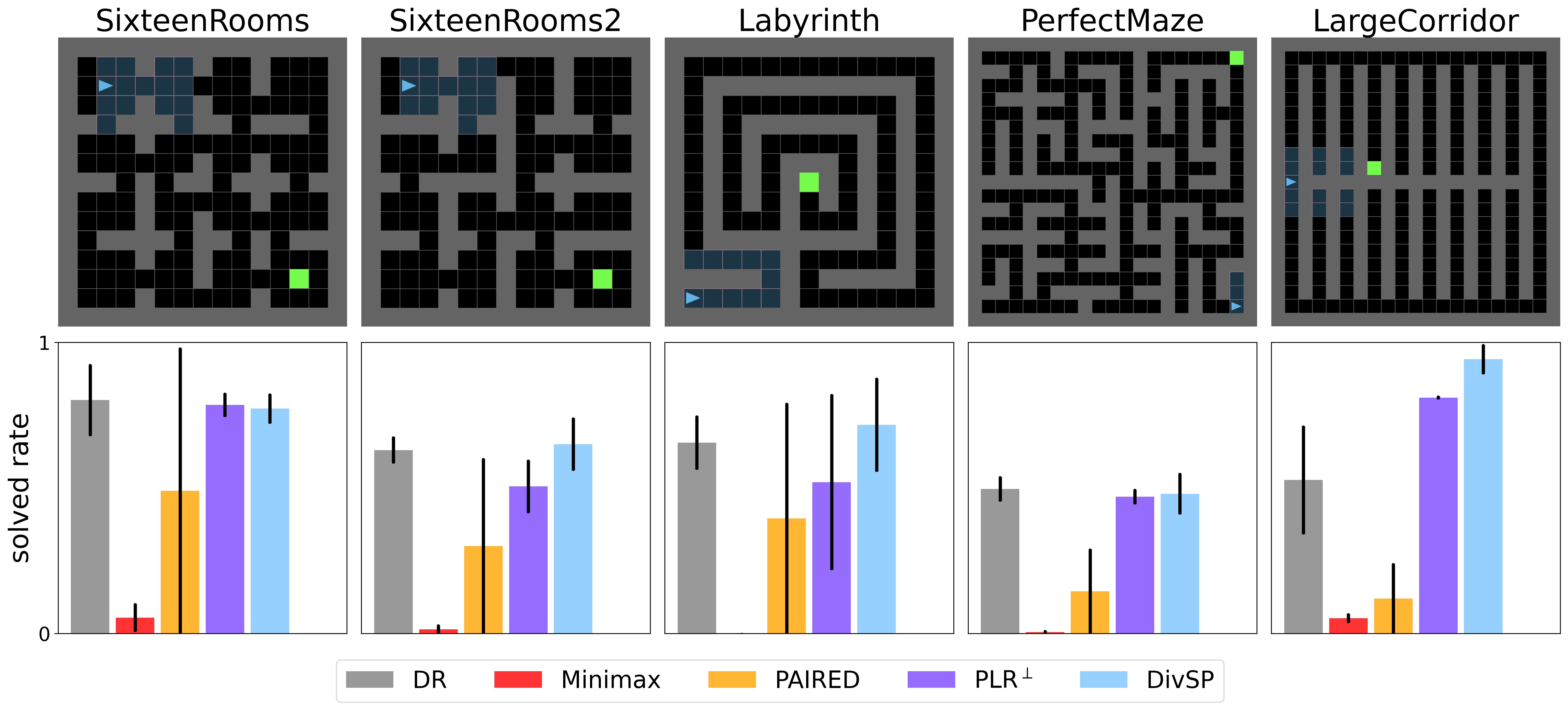}
    \caption{Zero-shot transfer performances in challenging environments after 100 million training steps. We show the median and interquartile range of solved rates over 5 runs.}
    \label{fig:maze_result}
\end{figure*}

\begin{itemize}[leftmargin=*]
    \item (\textbf{Generating new level}): 
    When there is a newly generated level, we measure the diversity score of the new level and the levels in the level buffer according to Equation~\ref{eq:def_div}. To achieve a diverse level buffer, if the diversity score of a new level is lower than one of the levels in the buffer, we add the new level $\theta_{new}$ to the buffer $\Lambda$ to replace the level with the highest diversity score $F_{div}(,)$ (Line 11 and 12). Note that, unlike the state representative where we want the diversity score to be high because we hope every observation collected from the level can find a similar observed state in the set $S_{env}$; instead, here we want a diverse level buffer, which implies minimal similarity. Therefore, the lower its diversity score, the less likely it is to find a similar level in the level buffer and be replaced.

    \item (\textbf{Sampling level from buffer}): To decide at which level to train, we assign each level $\theta_i$ a probability that is based on the combination of its diversity score and learning potential. Following~\citet{jiang2021prioritized}, we employ the GAE function shown in Equation~\ref{eq:gae} as the proxy for its learning potential. Given the learning potential of $F_{gae}(\theta_i)$, we first rank them accordingly and then use a prioritization function $h$ to define how differences in learning potential translated into differences in prioritization. As a result, we can get a learning potential prioritized distribution $P_{gae}(\Lambda)$ over the level buffer, and the probability for $\theta_i$ is
    \begin{equation}
        P_{gae}(\theta_i|\Lambda, F_{gae}) = \frac{h\left( \text{rank}(F_{gae}(\theta_i)) \right)^{1/\beta}}{\sum_j h\left(\text{rank}(F_{gae}(\theta_j)) \right)^{1/\beta}}
    \end{equation}
    where $\beta$ is the temperature parameter that tunes how much $h(\text{rank}(F_{gae}(\theta_i)))$ determines related probability, and $\text{rank}(F_{gae}(\theta_i))$ is the rank of level $\theta_i$ sorted in the descending order among the level buffer. Same to~\citet{jiang2021prioritized}, we use $h\left(\text{rank}(F_{gae}(\theta_j)) \right) = \frac{1}{\text{rank}(F_{gae}(\theta_i))}$. Similarly, we can compute the diversity score computed through Equation~\ref{eq:def_div} and its corresponding diversity score prioritized distribution $P_{div}(\Lambda)$ over the level buffer~\footnote{We rank $F_{div}(,)$ in the ascending order.}. Combining on $P_{gae}(\Lambda)$, based on the learning potential, and $P_{div}(\Lambda)$, based on diversity score, we update the overall replay distribution $P_{replay}(\Lambda) $ over $\Lambda$ as (Line 19):
    \begin{equation}\label{eq:replay_prob}
        P_{replay}(\Lambda) = (1-\rho)\cdot P_{gae}(\Lambda) + \rho\cdot P_{div}(\Lambda)
    \end{equation}
    Therefore, when a level has a higher learning potential or is significantly different from other levels in the level buffer, it has a higher chance of being sampled. We present the overall framework for \algo in Figure~\ref{fig:framework}.
\end{itemize}

% We present the overall framework for \algo in Figure~\ref{fig:framework}

% \begin{figure*}[t]
%     \centering
%     \includegraphics[width=1.0\linewidth]{walker_result.png}
%     \caption{Performance on test environments during training (mean and standard error).}
%     \label{fig:walker_result}
% \end{figure*}

% 

\begin{figure}[t]
    \centering
    \includegraphics[width=1.0\linewidth]{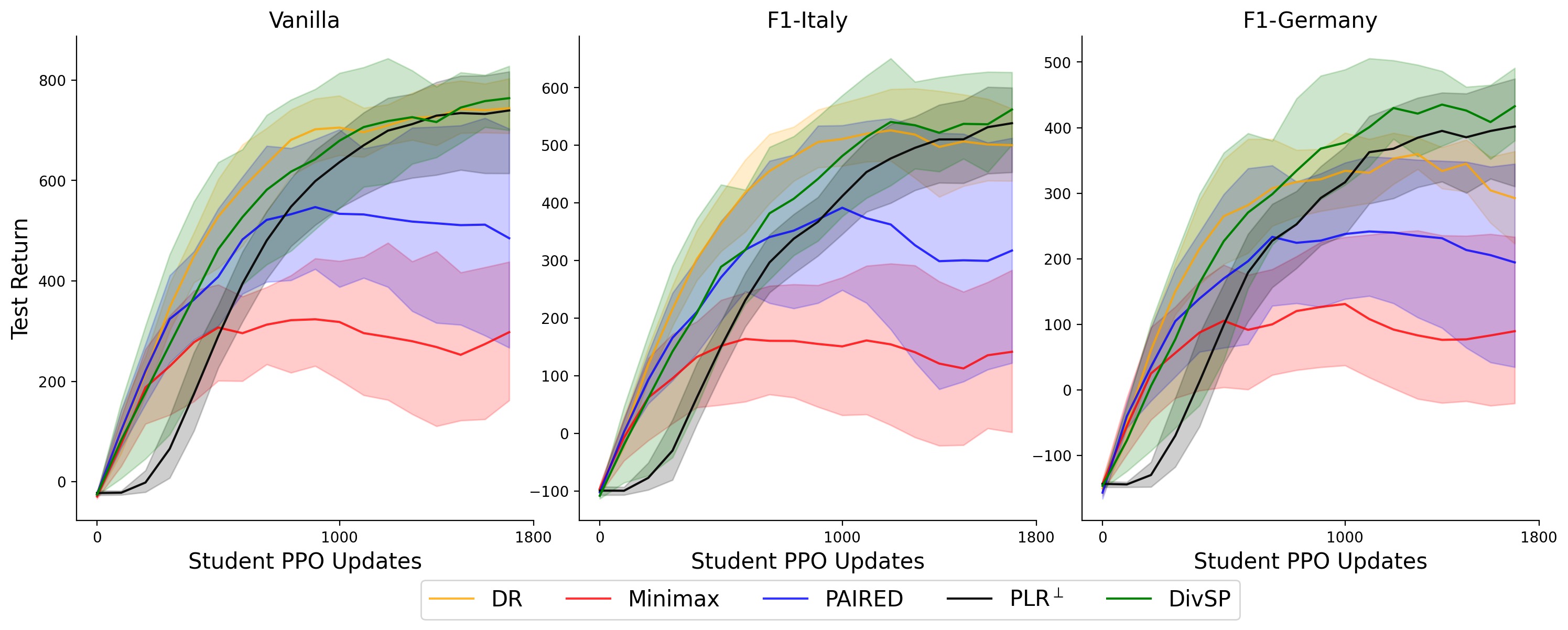}
    \caption{Zero-shot transfer performance on the OOD F1 tracks: Vanilla, Italy and Germany. We provide mean and standard deviation over 5 runs.}
    \label{fig:car_result}
\end{figure}

\section{Experiment}
In this section, we present our experimental results in the domains of \emph{BipedalWalker}, \emph{Minigrid}, and \emph{CarRacing} to demonstrate and illustrate the outperformance of our approach when a trained agent is transferred to new environments. We compare our approach against existing UED methods: Domain Randomization (DR~\cite{tobin2017domain}), Minimax~\cite{wang2019paired}, PAIRED~\cite{dennis2020emergent}, PLR~\cite{jiang2021replay}. 
We show the average and variance of the performance for our method, baselines with five random seeds.

\subsection{Performance on BipedalWalker}
\begin{figure}[t]
    \centering
    \includegraphics[width=1.0\linewidth]{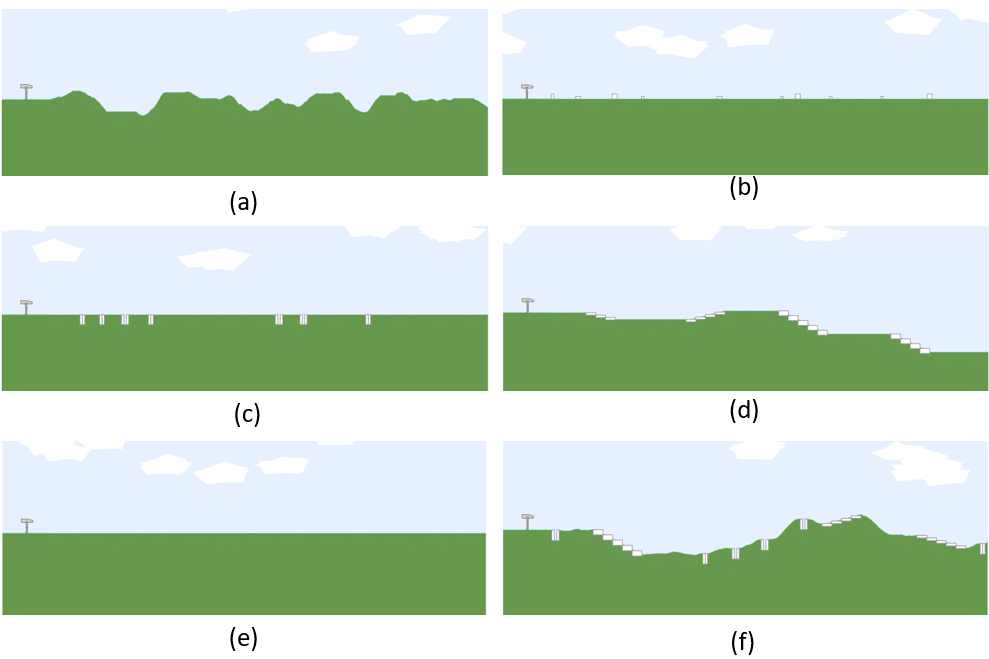}
    \caption{An illustration of levels generated with four kinds of obstacles: (a) Roughness of range (2,8) (b) Stump height of range (1,3) (c) Pit gap of range(1,3) (d) Stair steps of range (2,6) (e) Vanilla \emph{BipedalWalker} (f)Hard \emph{BipedalWalker} with a mix of (a) to (d) parameters.}
    \label{fig:walker_example}
\end{figure}

We first evaluate our approach on \emph{BipedalWalker}~\cite{wang2019paired} environment. This environment entails continuous control with dense rewards. Similar to~\citet{wang2019paired}, we use a modified version of \emph{BipedalWalker-Hardcore} from OpenAI Gym~\cite{brockman2016openai}. In \emph{BipedalWalker}, there are 8 parameters that indirectly represent the intensity of four kinds of terrain-based obstacles for a two-legged robot: the minimum/maximum roughness of the ground, the minimum/maximum height of stump obstacles, the minimum/maximum width of pit gap obstacles, and the minimum/maximum size of ascending and descending flights of stairs. We provide an illustration of these four kinds of obstacles in Figure~\ref{fig:walker_example}.

The agent receives a 24-dimensional proprioceptive state with respect to its lidar sensors, angles, and contacts. The action space consists of four continuous values that control the torques of its four motors. In this environment, the teacher learns to control 8 parameters corresponding to the range of 4 kinds of obstacles and then combines a random seed to generate a specific level. All agents are trained with Proximal Policy Optimization (PPO,~\cite{schulman2017proximal}). 
% We train our student agent for 10k update (which is equivalent to around )
For a fair comparison, during training, we presented a vanilla \emph{BipedalWalker}, a challenging \emph{BipedalWalker-Hardcore} environment, and four specific levels in the context of isolated challenges in \{Roughness, Stump height, Pit gap, Stair step\} to evaluate our algorithm and baselines.

Figure~\ref{fig:walker_result} shows the transfer performance throughout training. As shown in the figure, \algo outperforms all baselines in most test environments, achieving a faster convergence. Those results support the key drivers of \algo's motivation: Producing incremental challenging environments at the frontier of the agent's capabilities and maintaining diverse environments to improve sample efficiency and build effective exploration. 

\subsection{Performance on Minigrid}
\begin{figure}[ht]
    \centering
    \includegraphics[width=0.95\linewidth]{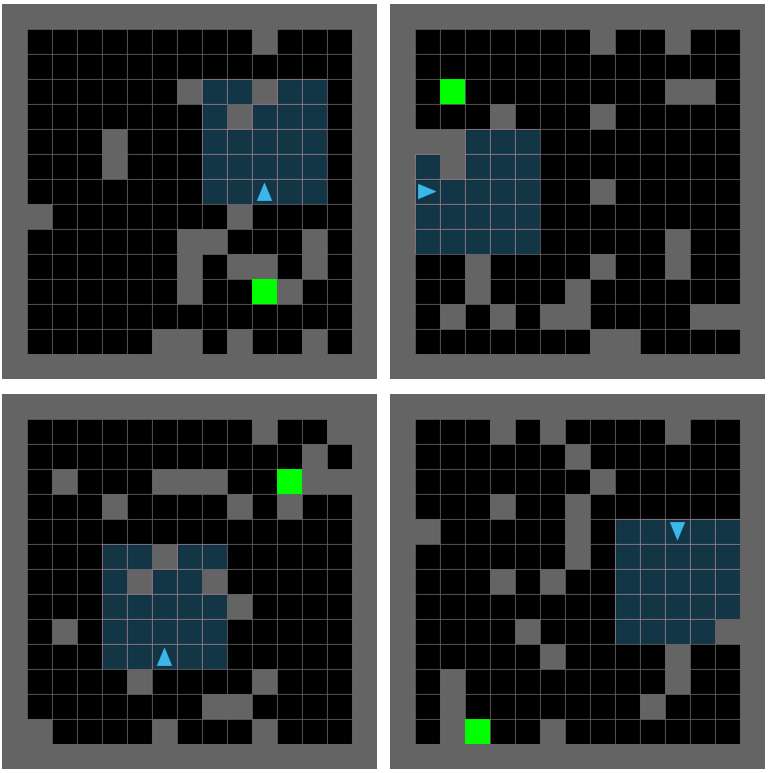}
    \caption{Example levels generated by \algo during training. The generator can place up to 25 blocks.}
    \label{fig:maze_example}
\end{figure}

Here we investigate the maze navigation environment introduced by \citet{dennis2020emergent}, which is based on \emph{Minigrid}~\cite{chevalier2018minimalistic}. We train the environment generator to learn how to build maze environments by choosing the location of the obstacles, the goals, and the starting location of the agent. Specifically, at the beginning of each iteration, the generator will place the student agent and the goal, and then every time step afterward, the generator outputs a location where the next obstacle will be placed. There will be up to 25 blocks that can be placed. We give several examples of generated mazes during the training in Figure~\ref{fig:maze_example}.

The maze environment is partially observable, where the student agent's view is shown as a blue-shaded area in Figure~\ref{fig:maze_example}. The student agent (blue triangle) must explore the maze to find a goal (green square). In order to deal with the partially observable setting, our agents use PPO with a Recurrent Neural Network structure.
We compare our agents' transfer ability trained by different approaches on human-designed levels. The test environments and the performance are reported in Figure~\ref{fig:maze_result}. While DR acts as a strong baseline in this domain, \algo can attain the similar highest mean return. 
% Furthermore, our RL environment generator can successfully learn to maximize the regret and generate the solvable levels through the self-play.

% \begin{figure*}[t]
%     \centering
%     \includegraphics[width=0.8\linewidth]{maze_result.png}
%     \caption{Zero-shot transfer performances in challenging environments after 100 million training steps. We show the median and interquartile range of solved rates over 5 runs.}
%     \label{fig:maze_result}
% \end{figure*}

\begin{table*}[t]
\caption{The components of baselines. Like PAIRED, our algorithm \algo  uses regret to train the generator, but it also replay levels according to its regret value and diversity score. Here MCC is an abbreviation for Minimal Criteria Coevolution~\cite{wang2019paired}.
% Both Domain Randomization and PLR randomly generate level to train the agent.
}
\centering
\begin{tabular}{|c|c|c|c|c|}
\hline
Algorithm      & Generation Strategy & Replay Strategy &Buffer Objective & Setting   \\ \midrule
DR~\cite{tobin2017domain}      & Random           & None  &None & Single Agent \\ 
Minimax~\cite{wang2019paired}  &Evolution & MCC  & Minimax & Population-Based  \\ 
PAIRED~\cite{dennis2020emergent} & RL        & None & None & Single Agent  \\ 
PLR~\cite{jiang2021prioritized,jiang2021replay} &Random & Minimax Regret &Learning potential & Single Agent \\ \midrule
\algo    & RL & Max Regret and Diversity &Diversity& Single Agent  \\ \bottomrule
\end{tabular}
\label{tab:baseline}
\end{table*}

\subsection{Performance on CarRacing}
\begin{figure}[t]
    \centering
    \includegraphics[width=1.0\linewidth]{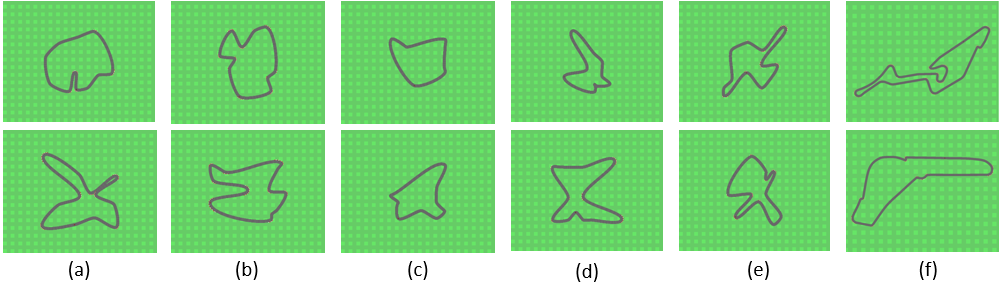}
    \caption{A randomly-selected examples of \emph{CarRacing} tracks produced by different algorithms. (a)DR  (b) Minimax (c) PAIRED (d) PLR (e)\algo   (f) Two examples in the \emph{CarRacing} F1 benchmark that are used for evaluating zero-shot generalization.}
    \label{fig:car_example}
\end{figure}

Finally, we investigate the learning dynamics of \algo  and baselines on \emph{CarRacing}~\cite{brockman2016openai}, a popular continuous-control environment with dense rewards. Similar to the partially observable navigation task, The student agent in \emph{CarRacing} receives a partial, pixel observation and has a 3-dimensional action space. The goal of the agent is to drive a full lap around a generated track. To generate a feasible level (closed-loop track), following~\cite{jiang2021replay}, the generator learns to choose a sequence of up to 12 control points, which will unique generate a B\'{e}zier curve~\cite{mortenson1999mathematics} within predefined curvature constraints. In Figure~\ref{fig:car_example}, we show some examples of \emph{CarRacing} tracks produced by different algorithms.

We present per-track zero-shot transfer returns of policies trained by each method on some of the human-designed Formula One (F1) tracks throughout training in Figure~\ref{fig:car_result}. Note that these tracks are significantly out-of-distribution (OOD) as they can not be generated within 12 control points. Remarkably, \algo can either mitigate the degeneracy of PAIRED or achieve significant outperformance than other baselines in mean performance, providing further evidence of the benefits of the induced curriculum and diverse level buffer.

% \begin{figure}[t]
%     \centering
%     \includegraphics[width=1.0\linewidth]{car_result.png}
%     \caption{Zero-shot transfer performance on the OOD F1 tracks: Vanilla, Italy and Germany. We provide mean and standard deviation over 5 runs.}
%     \label{fig:car_result}
% \end{figure}

\section{Related Work}
% \begin{table*}[t]
% \caption{The components of baselines. Like PAIRED, our algorithm \algo  uses regret to train the generator, but it also replay levels according to its regret value and diversity score. Here MCC is an abbreviation for Minimal Criteria Coevolution~\cite{wang2019paired}.
% % Both Domain Randomization and PLR randomly generate level to train the agent.
% }
% \centering
% \begin{tabular}{|c|c|c|c|c|}
% \hline
% Algorithm      & Generation Strategy & Replay Strategy &Buffer Objective & Setting   \\ \midrule
% DR~\cite{tobin2017domain}      & Random           & None  &None & Single Agent \\ 
% Minimax~\cite{wang2019paired}  &Evolution & MCC  & Minimax & Population-Based  \\ 
% PAIRED~\cite{dennis2020emergent} & RL        & None & None & Single Agent  \\ 
% PLR~\cite{jiang2021prioritized,jiang2021replay} &Random & Minimax Regret &Learning potential & Single Agent \\ \midrule
% \algo    & RL & Max Regret and Diversity &Diversity& Single Agent  \\ \bottomrule
% \end{tabular}
% \label{tab:baseline}
% \end{table*}

This work aims to train agents that are capable of generalizing across a wide range of environments~\cite{whiteson2009generalized}. Several methods for enhancing generalization in RL use techniques from supervised learning, such as data augmentation~\cite{raileanu2020automatic,kostrikov2020image,wang2020improving}, and feature distillation~\cite{igl2020impact}. 
Opposed to supervised learning, there is a trend of introducing the mechanism of curricula in different learning situations~\cite{fang2019curriculum,weinshall2020theory,wu2020curricula}. In RL, curricula improve the learning performance of the agent by adapting the training environment to the agent's current capabilities.
One prior approach is domain randomization~\cite{jakobi1997evolutionary,tobin2017domain}, where they train the agent to a wide range of randomly generated environments. In contrast, \citet{akkaya2019solving} propose automatic domain randomization, where they use a curriculum that gradually increases the difficulty for agent training. 
In the multi-task domain~\cite{sukhbaatar2017intrinsic,zhang2020automatic,du2022takes,klink2022curriculum}, they particularly set up an automatic curriculum for goals that agent needs to solve. Curricula are often generated as the proposed goals are right at the frontier of the learning process of an agent.

In particular, we focus on a growing corpus of work in unsupervised environment design~\cite{dennis2020emergent}, which is inherently related to the Automatic Curriculum Learning~\cite{florensa2017reverse,portelas2020automatic}. It seeks to learn a curriculum of adaptively generating challenging environments to train robust agents. \citet{dennis2020emergent} proposed PAIRED algorithm, where they introduce an environmental adversary that learns a curriculum to control environmental parameters to maximize approximate regret. POET~\cite{wang2019paired, wang2020improving} co-evolves the generation of environmental challenges and the optimization of agents to solve them. \cite{jiang2021prioritized,jiang2021replay} further introduce PLR, a general framework where agents can revisit previously generated environments with high learning potential for training. We take inspiration from the automatically generated environments with a curriculum design and maintain a level buffer with high learning potential and high diversity.

For brevity, we provide the works most related to our approach and summarize them in Table~\ref{tab:baseline}.

\section{Conclusion}
In this paper, we introduce \algo, a novel method for unsupervised environment design, that evolves a curriculum to automatically create a distribution of training environments through self-play. Furthermore, in order to improve sample efficiency, \algo selectively revisit previously generated environments by prioritizing those with higher estimated learning potential and diversity. Finally, we performed experiments in various benchmark environments and demonstrated that our algorithm leads to superior zero-shot transfer performance in most of the settings. 

\bibliography{example_paper}
\bibliographystyle{plainnat}

\appendix

\end{document}